\documentclass[conference]{IEEEtran}
\usepackage{times}

\usepackage[numbers]{natbib}
\usepackage{amsmath,amssymb,amsthm,bm}
\usepackage{multicol}
\usepackage[bookmarks=true]{hyperref}
\usepackage{graphicx}
\usepackage{graphics} 
\usepackage[tight]{subfigure}
\usepackage[ruled, linesnumbered, vlined]{algorithm2e}
\usepackage[group-separator={,}, group-minimum-digits=3]{siunitx}
\usepackage{tabularx,pbox,booktabs,caption}

\begin{document}

\title{\textbf{CALI}: \textbf{C}oarse-to-Fine \textbf{ALI}gnments Based Unsupervised Domain Adaptation of Traversability Prediction for Deployable Autonomous Navigation}



\author{\authorblockN{Zheng Chen}
\authorblockA{Luddy School of Informatics,\\ Computing, and Engineering \\ Indiana University - Bloomington\\
Indiana 47408\\
Email: zc11@iu.edu}
\and
\authorblockN{Durgakant Pushp}
\authorblockA{Luddy School of Informatics,\\ Computing, and Engineering \\ Indiana University -  Bloomington\\
Indiana 47408\\
Email: dpushp@iu.edu}
\and
\authorblockN{Lantao Liu}
\authorblockA{Luddy School of Informatics,\\ Computing, and Engineering \\ Indiana University -  Bloomington\\
Indiana 47408\\
Email: lantao@iu.edu}
}


%

\maketitle

\begin{abstract}
Traversability prediction is a fundamental perception capability for autonomous navigation. The diversity of data in different domains imposes significant gaps to the prediction performance of the perception model. 
In this work, we make efforts to reduce the gaps by proposing a novel coarse-to-fine \textit{unsupervised domain adaptation} (UDA) model -- CALI. Our aim is to transfer the perception model with high data efficiency, eliminate the prohibitively expensive data labeling, and improve the generalization capability during the adaptation from easy-to-obtain \textit{source domains} to various challenging \textit{target domains}. 
We prove that a combination of a coarse alignment and a fine alignment can be beneficial to each other and further design a first-coarse-then-fine alignment process. 
This proposed work bridges theoretical analyses and algorithm designs, leading to an efficient UDA model with easy and stable training. We show the advantages of our proposed model over multiple baselines in several challenging domain adaptation setups. To further validate the effectiveness of our model, we then combine our perception model with a visual planner to build a navigation system and show the high reliability of our model in complex natural environments where no labeled data is available.
\footnote{The paper is published in Robotics: Science and Systems (RSS) 2022.}\\
The robot navigation demonstration can be seen in this video:  \href{https://www.youtube.com/watch?v=Nqsegaq_x-o}{\url{https://www.youtube.com/watch?v=Nqsegaq_x-o}}.
\end{abstract}

\IEEEpeerreviewmaketitle

\section{Introduction}
\label{sec:introduction}


We consider the deployment of autonomous robots in the real-world unstructured field environments, where the environments can be extremely complex involving random obstacles (e.g., big rocks, tree stumps, man-made objects), cross-domain terrains (e.g., combinations of gravel, sand, wet, uneven surfaces), as well as dense vegetation (tall and low grasses, shrubs, trees). Whenever a robot is deployed in such an environment, it needs to understand which area of the captured scene is navigable. A typical solution to this problem is the visual traversability prediction that can be achieved by learning the {\em scene semantic segmentation}. 


Visual traversability prediction has been tackled by using deep neural networks where the models are typically trained offline with well-labeled datasets in limited scenarios. However, there is a gap between the data used to train the model and the real world. It is usually challenging for existing datasets to well approximate the true distributions of the unseen {\em target} environments where the robot is deployed. Even incrementally collecting and adding new training data on the fly cannot guarantee the target environments to be well {\em in-distribution} included. In addition, annotating labels for dense predictions, e.g., semantic segmentation, is prohibitively expensive. Therefore, developing a generalization-aware deep model is crucial for 
robotic systems considering the demands of the practical deployment of deep perception models and the costs/limits of collecting new data in many robotic applications, e.g., autonomous driving, search and rescue, and environmental monitoring.

\begin{figure}
{
\centering
  {\includegraphics[width=0.98\linewidth]{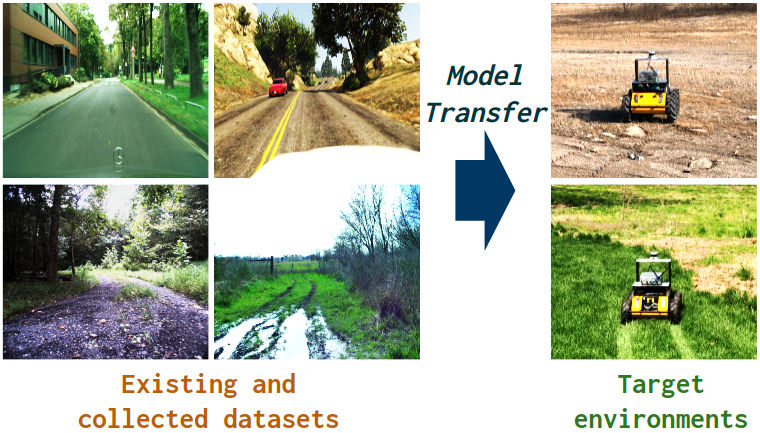} 
  } 
\caption{\small Transferring models from the available domain to the target domain. 
The existing available data might be from either a simulator or collecting data in certain environments, at a certain time, and with certain sensors. In contrast, the target deployment might have significantly varying environments, time, 
and sensors.}  \vspace{-20pt}
\label{fig:intro} 
}
\end{figure}



To tackle this challenge, a broadly studied framework is \textit{transfer learning}~\cite{pan2009survey} which aims to transfer models between two domains -- \textit{source domain} and \textit{target domain} -- that have related but different data distributions. The prediction on target domain can be considered as a strong generalization since testing data (in target domain) might fall out of the {independently and identically distributed (i.i.d.)}  assumption and follow a very different distribution than the training data (in source domain). The ``transfer"  process has significant meaning to our model development since we can view the available public datasets~\cite{richter2016playing, Cordts2016Cityscapes, RUGD2019IROS, jiang2020rellis3d} as the source domain and treat the data in the to-be-deployed environments as the target domain. In this case, we have access to images and corresponding labels in source domain and images in target domain, but no access to labels in target domain. Transferring models, in this set-up, is called  \textit{Unsupervised Domain Adaptation}~(UDA)~\cite{wilson2020survey, zhang2021survey}.


Domain Alignment (DA)~\cite{ganin2016domain, hoffman2016fcns, hoffman2018cycada, tsai2018learning, vu2019advent} and Class Alignment (CA) \cite{saito2018maximum} are two conventional ways to tackle the UDA problem. DA treats the deep features as a whole. It works well for image-level tasks such as image classification, but has issues with pixel-level tasks such as semantic segmentation, as the alignment of whole distributions ignores the class features and might misalign class distributions, even the whole features from the source domain and target domain are already well-aligned. CA is proposed to solve this issue for dense predictions with multiple classes.

It is natural and necessary to use CA to tackle the UDA of semantic segmentation as we need to consider aligning class features. However, CA can be problematic and might fail to outperform the DA for segmentation, and in a worse case, might have unacceptable \textit{negative transfer}, which means the performance with adaptation is even degraded than that without adaptation. Intuitively, we need to consider more alignments in CA than in DA. Thus the searching space might be more complicated, and training might be more unstable and harder to converge to an expected minima, leading to larger prediction errors.

To solve the issue of CA, we investigate the relationship of the upper bounds of the prediction error on target domain between DA and CA and provide a theoretical analysis of the upper bounds of target prediction error in UDA setup, and bridge the theoretical analysis and algorithm design for UDA of traversability prediction.

In summary, our contributions include
\begin{itemize}
    \item 
    We prove that with proper assumptions, the upper bound of CA is upper bounded by the upper bound of DA. This indicates that constraining the training of CA using DA can be beneficial. 
    We then propose a novel concept of \textit{pseudo-trilateral game structure}~(PTGS) for integrating DA and CA.
    \item
    We propose an efficient coarse-to-fine alignments based UDA model, named CALI, for traversability prediction. The new proposal includes a trilateral network structure, novel training losses, and an alternative training process. Our model design is well supported by theoretical analysis. It is also easy and stable to train and converge.
    \item
    We show significant advantages of our proposed model compared to several baselines in multiple challenging public datasets and one self-collected dataset. 
    We combine the proposed segmentation model and a visual planner to build a visual navigation system. The results show high safety and effectiveness of our model.
\end{itemize}



\section{Related Work}
\label{sec:related work}
\textbf{Semantic Segmentation:} Semantic segmentation aims to predict a unique human-defined semantic class for each pixel in the given images. With the prosperity of deep neural networks, the performance of semantic segmentation has been boosted significantly, especially by the advent of FCN~\cite{long2015fully} that first proposes to use deep convolutional neural nets to predict segmentation. Following works try to improve the FCN performance by multiple proposals, e.g., using different sizes of kernels or dilation rates to aggregate multi-scale features \cite{chen2017deeplab, chen2017rethinking, yu2015multi}; building image pyramids to create multi-resolution inputs~\cite{zhao2017pyramid}; applying probabilistic graph to smooth the prediction~\cite{liu2017deep}; compensating features in deeper level by an encoder-decoder structure~\cite{ronneberger2015u}, and employing attention mechanism to capture the long-range dependencies among pixels in a more compact and efficient way~\cite{ranftl2021vision}. We can also see how excellent the current semantic segmentation SOTA performance is from very recently released work~\cite{zheng2021rethinking, xie2021segformer}. However, all of those methods belong to fully-supervised learning and the performance might catastrophically be degraded when a domain shift exists between the training data and the data when deploying. Considering the possible domain shift and developing \textit{adaptation-aware} models is extremely practical and urgent. 

\textbf{Unsupervised Domain Adaptation:} The main approaches to tackle UDA include adversarial training (a.k.a., \textit{distribution alignment})~\cite{ganin2016domain, hoffman2016fcns, hoffman2018cycada, tsai2018learning, saito2018maximum, vu2019advent, luo2019taking, wang2020classes} and self-training~\cite{zou2018unsupervised, zhang2017curriculum, mei2020instance, hoyer2021daformer}. Although self-training is becoming another dominant method for segmentation UDA in terms of the empirical results, it still lacks a sound theoretical foundation. In this paper, we only focus on the alignment-based methods that not only keep close to the UDA state-of-the-art (SOTA) performance but also are well supported by sound theoretical analyses~\cite{ben2007analysis, blitzer2008learning, ben2010theory}. 

The alignment-based methods adapt models via aligning the distributions from the source domain and target domain in an adversarial training process, i.e., making the deep features of source images and target images indistinguishable to a discriminator net. Typical approaches to UDA include Domain Alignment (DA) \cite{ganin2016domain, hoffman2016fcns, hoffman2018cycada, tsai2018learning, vu2019advent}, which aligns the two domains using global features (aligning the feature tensor from source or target \textit{as a whole}) and Class Alignment (CA) \cite{saito2018maximum, luo2019taking, wang2020classes}, which only considers aligning features of each class from source and target, no matter whether the domain distributions are aligned or not. In \cite{saito2018maximum}, the authors are inspired by the theoretical analysis of \cite{ben2010theory} and propose a discrepancy-based model for aligning class features. There is a clear relation between the theory guidance \cite{ben2010theory} and the design of network, loss, and training methods. There are some recent works \cite{luo2019taking, wang2020classes} similar to the proposed work in spirit and show improved results compared to \cite{saito2018maximum}, but it is still unclear to relate the proposed algorithms with theory and to understand why the structure/loss/training is designed as the presented way.

\section{Background and Preliminary Materials}
We consider segmentation tasks where the input space is $\mathcal{X}\subset \mathbb{R}^{H\times W\times 3}$, representing the input RGB images, and the label space is $\mathcal{Y}\subset \left\{ 0, 1\right\}^{H\times W\times K}$, representing the ground-truth $K$-class segmentation images, where the label for a single pixel at $(h, w)$ is denoted by a one-hot vector $y^{(h, w)}\in \mathbb{R}^K$ whose elements are by-default 0-valued 
except at location $(h, w)$ which is labeled as 1. Domain adaptation has two domain distributions over $\mathcal{X}\times \mathcal{Y}$, named source domain $\mathcal{D}_S$ and target domain $\mathcal{D}_T$. 
In the setting of UDA for segmentation, we have access to $m_s$ \textit{i.i.d.} samples with labels $\mathcal{U}_S=\left\{\mathbf{x}_{si}, \mathbf{y}_{si} \right\}_{i=1}^{m_s}$ from $\mathcal{D}_S$ and $m_t$ \textit{i.i.d.} samples without labels $\mathcal{U}_T=\left\{\mathbf{x}_{tj} \right\}_{j=1}^{m_t}$ from $\mathcal{D}_T$. 

In the UDA problem, we need to reduce the prediction error on the target domain. A {\em hypothesis} is a function $h: \mathcal{X}\rightarrow \mathcal{Y}$. We denote the space of $h$ as $\mathcal{H}$. With the loss function $l(\cdot, \cdot)$, the expected error of $h$ on $\mathcal{D}_S$ is defined as
\begin{equation}
    \label{eq:source_expected_error}
    \epsilon_S(h):=\mathbb{E}_{(x, y)\sim \mathcal{D}_S}l(h(x), y).
\end{equation}
Similarly, we can define the expected error 
of $h$ on $\mathcal{D}_T$ as
\begin{equation}
    \label{eq:target_expected_error}
    \epsilon_T(h):=\mathbb{E}_{(x, y)\sim \mathcal{D}_T}l(h(x), y).
\end{equation}

Two important upper bounds related to the source and target error are given in \cite{ben2010theory}. Basically, 

\textbf{Theorem 1} 
For a hypothesis $h$, \begin{equation}
    \label{eq:theorem1}
    \epsilon_T(h)\leq \epsilon_S(h)+d_1(\mathcal{D}_S, \mathcal{D}_T)+\lambda,
\end{equation}
where $d_1(\cdot, \cdot)$ is the $L^1$ divergence for two distributions, and the constant term $\lambda$ does not depend on any $h$. 
However, it is claimed in \cite{ben2010theory} that the bound with $L^1$ divergence cannot be accurately estimated from finite samples, and using $L^1$ divergence can unnecessarily inflate the bound. Another divergence measure is thus introduced to replace the $L^1$ divergence with a new bound derived.

\textbf{Definition 1} 
Given two domain distributions $\mathcal{D}_S$ and $\mathcal{D}_T$ over $\mathcal{X}$, and a hypothesis space $\mathcal{H}$ that has finite VC dimension, the $\mathcal{H}$-divergence between $\mathcal{D}_S$ and $\mathcal{D}_T$ is defined as
\begin{equation}
    \label{eq:h_divergence}
    \begin{aligned}
        d_{\mathcal{H}}(\mathcal{D}_S, \mathcal{D}_T) = 2\sup_{h\in \mathcal{H}} | &\text{P}_{x\sim \mathcal{D}_S} \left [ h(x)=1 \right ] - \\
        &\text{P}_{x\sim \mathcal{D}_T} \left [ h(x)=1 \right ] |,
    \end{aligned}
\end{equation}
where $\text{P}_{x\sim \mathcal{D}_S}[h(x)=1]$ represents the probability of $x$ belonging to $\mathcal{D}_S$. Same to $\text{P}_{x\sim \mathcal{D}_T} \left [ h(x)=1 \right ]$.

The $\mathcal{H}$-divergence resolves the issues in the $L^1$ divergence. If we replace  $d_1(\mathcal{D}_S, \mathcal{D}_T)$ in Eq.~\eqref{eq:theorem1} with $d_{\mathcal{H}}(\mathcal{D}_S, \mathcal{D}_T)$, then a new upper bound for $\epsilon_T(h)$, named as $\mathbb{UB}_1$,  can be written as
\begin{equation}
    \label{eq:ub1}
    \begin{aligned}
        &\epsilon_T(h)\leq \mathbb{UB}_1,\\
        & \mathbb{UB}_1 = \epsilon_S(h)+d_{\mathcal{H}}(\mathcal{D}_S, \mathcal{D}_T) + \lambda.
    \end{aligned}
\end{equation}

An approach to compute the empirical $\mathcal{H}$-divergence is also proposed in \cite{ben2010theory}.

\textbf{Lemma 1} 
For a symmetric hypothesis class $\mathcal{H}$ (one where for every $h\in \mathcal{H}$, the inverse hypothesis $1-h$ is also in $\mathcal{H}$) and two sample sets $\mathcal{U}_S=\left\{ x_i, i=1, \cdots, m_s, x_i\sim \mathcal{D}_S\right\}$ and $\mathcal{U}_T=\left\{ x_j, j=1, \cdots, m_t, x_j\sim \mathcal{D}_T\right\}$.
\begin{equation}
    \label{eq:empirical_h_divergence}\
    \begin{aligned}
        \hat{d}_{\mathcal{H}}(\mathcal{D}_S, \mathcal{D}_T) = 2 \Bigg( \Bigg.  1-\min_{\eta \in \mathcal{H}} \Bigg[ \Bigg. &\frac{1}{m_s} \sum_{i=1}^{m_s} \mathbb{I}[\eta(x_i)=0] +\\
        &\frac{1}{m_t}\sum_{j=1}^{m_t} \mathbb{I}[\eta(x_j)=1] \Bigg. \Bigg] \Bigg. \Bigg),
    \end{aligned}
\end{equation}
where $\mathbb{I}[a]$ is an indicator function which is 1 if $a$ is true, and $0$ otherwise.

The second upper bound is based on a new hypothesis called symmetric difference hypothesis.

\textbf{Definition 2} 
For a hypothesis space $\mathcal{H}$, the symmetric difference hypothesis space $\mathcal{H}\Delta \mathcal{H}$ is the set of hypotheses
\begin{equation}
    \label{eq:hdh_space}
    g\in \mathcal{H}\Delta\mathcal{H} \Leftrightarrow g(x) = h(x)\oplus  h'(x)~~~\text{for some } h, h'\in \mathcal{H},
\end{equation}
where $\oplus$ denotes an XOR operation. Then we can define the $\mathcal{H}\Delta\mathcal{H}$-distance as
\begin{equation}
    \label{eq:hdh_distance}
    \begin{aligned}
    d_{\mathcal{H}\Delta\mathcal{H}}(\mathcal{D}_S, \mathcal{D}_T)=2\sup_{h, h'\in \mathcal{H}} | &\text{P}_{x\sim \mathcal{D}_S} \left [ h(x)\neq h'(x) \right ] - \\
        &\text{P}_{x\sim \mathcal{D}_T} \left [ h(x)\neq h'(x) \right ] |.
    \end{aligned}
\end{equation}
Similar to Eq.~(\ref{eq:ub1}), if we replace $d_1(\mathcal{D}_S, \mathcal{D}_T)$ with $d_{\mathcal{H}\Delta\mathcal{H}}(\mathcal{D}_S, \mathcal{D}_T)$,  the second upper bound for $\epsilon_T(h)$, named as $\mathbb{UB}_2$,  can be expressed as
\begin{equation}
    \label{eq:ub2}
    \begin{aligned}
    &\epsilon_T(h)\leq \mathbb{UB}_2,\\
    &\mathbb{UB}_2 = \epsilon_S(h)+d_{\mathcal{H}\Delta\mathcal{H}}(\mathcal{D}_S, \mathcal{D}_T) + \lambda,
    \end{aligned}
\end{equation}
where $\lambda$ is the same term as in Eq.~(\ref{eq:theorem1}).

A standard way to achieve the alignment for deep models is to use the adversarial training method, which is also used in Generative Adversarial Networks (GANs). Therefore we explain the key concepts of adversarial training using the example of GANs.

GAN is proposed to learn the distribution $p_r$ of a set of given data $\left\{ \mathbf{x} \right\}$ in an adversarial manner. The architecture consists of two networks - a generator $G$, and a discriminator $D$. The $G$ is responsible for generating fake data (with distribution $p_g$) from random noises $\mathbf{z}\sim p_{\mathbf{z}}$ to fool the discriminator $D$ that is instead to accurately distinguish between the fake data and the given data. Optimization of a GAN involves a mini-maximization over a joint loss for $G$ and $D$.
\begin{equation}
    \label{eq:gan}
    \begin{aligned}
        &\min_{G}\max_{D} V(G, D)\\
        &V(G, D) = \mathbb{E}_{\mathbf{x}\sim p_r} \log \left [ D(\mathbf{x}) \right ] + \mathbb{E}_{\mathbf{z}\sim p_{\mathbf{z}}}\log \left [ 1-D(G(z)) \right ].
    \end{aligned}
\end{equation}
where we use $1$ as the real label and $0$ as the fake label. Training with Eq.~(\ref{eq:gan}) is a bilateral game where the distribution $p_g$ is aligned with the distribution $p_r$.

The two bounds (Eq.~(\ref{eq:ub1}) and Eq.~(\ref{eq:ub2})) for the target domain error are separately given in \cite{ben2010theory}. It has been independently demonstrated that domain alignment corresponds to optimizing over $\mathbb{UB}_1$ \cite{ganin2016domain}, where optimization over the upper bound $\mathbb{UB}_1$ (Eq.~(\ref{eq:ub1}) with the divergence Eq.~(\ref{eq:empirical_h_divergence})) is proved as equivalent to Eq.~(\ref{eq:gan}) with a supervised learning in the source domain, and that class alignment corresponds to optimizing over $\mathbb{UB}_2$ \cite{saito2018maximum}, where the $d_{\mathcal{H}\Delta\mathcal{H}}$ is approximated by the discrepancy between two different classifiers.

Training DA is straightforward since we can easily define binary labels for each domain, e.g., we can use 1 as the source domain label and 0 as the target domain label. Adversarial training over the domain labels can achieve domain alignment. For CA, however, it is difficult to implement as we do not have target labels, hence the target class features are completely unknown to us, thus leading naively using adversarial training over each class impossible. The existing way well supported by theory to perform CA \cite{saito2018maximum} is to indirectly align class features by devising two different classifier hypotheses. The two classifiers have to be well trained on the source domain and are able to classify different classes in the source domain with different decision boundaries. Then considering the shift between source and target domain, the trained two classifiers might have disagreements on target domain classes. Note since the two classifiers are already well trained on the source domain, the agreements of the two classifiers represent those features in the target domain that are close to the source domain, while in contrast, the features where disagreements happen indicate that there is a large shift between source and target. We use the disagreements to approximate the distance between source and target. If we are able to minimize the disagreements of the two classifiers, then features of each class between source and target will be enforced to be well aligned.

\begin{figure}
{
\centering
  {\includegraphics[width=0.8\linewidth]{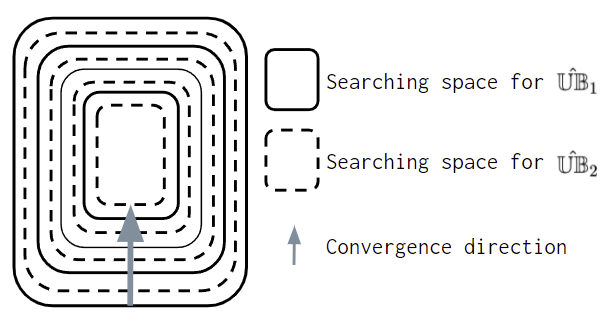} 
  } 
\caption{\small An ideal iterative training process by integrating DA and CA.}  \vspace{-10pt}
\label{fig:optim} 
}
\end{figure}

\section{Methodology}
\label{sec:methodology}

In this work we further investigate the relation between the $\mathbb{UB}_1$ and $\mathbb{UB}_2$ and prove that $\mathbb{UB}_1$ turns out to be an upper bound of $\mathbb{UB}_2$, meaning DA can be a necessary constraint to CA. This is also consistent to our intuition: DA aligns features globally in a coarse way while CA aligns features locally in a finer way. Constraining CA with DA is actually a coarse-to-fine process, which makes the alignment process efficient and stable. By carefully studying the internal structure of existing DA and CA work, we propose a novel concept, \textit{pseudo-trilateral game structure}, for efficiently integrating DA and CA. We follow our theoretical analysis and proposed PTGS to guide the development of CALI, including designs of model structure, losses and training process.

\begin{figure}
{
\centering
  {\includegraphics[width=0.7\linewidth]{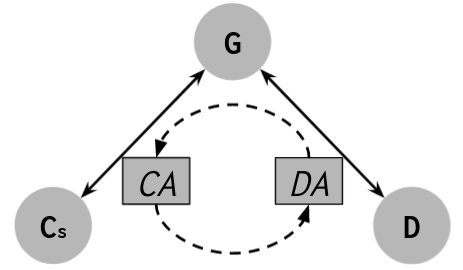} 
  } 
\caption{\small Pseudo-trilateral game structure (PTGS). Three players are in the game, a feature extractor $G$, a domain discriminator $D$ and a family of classifiers $C_{s}$. The game between $G$ and $Cs$ is the CA while the game between $G$ and $D$ is the DA. The DA and CA are connected by sharing the same feature extractor $G$. Both $D$ and $C_s$ are trying to adjust the $G$ such that the features between source and target generated from $G$ could be well aligned globally and locally.}  \vspace{-10pt}
\label{fig:trilateral} 
}
\end{figure}

Notations used in this paper is explained as follows. we denote the segmentation model $h$ as $h^{\theta, \phi}(x) = C^{\theta}(G^{\phi}(x))$ which consists of a feature extractor $G^{\phi}$ parameterized by $\phi$ and a classifier $C^{\theta}$ parameterized by $\theta$, and $x$ is a sample from $\mathcal{U}_S$ or $\mathcal{U}_T$. If multiple classifiers are used, we will denote the $j^{th}$ classifier as $C_{j}$. We denote the discriminator as $D^{\psi}$ parameterized by $\psi$.


\subsection{Bounds Relation}
\label{sec:bounds_relation}
We start by examining the relationship between the DA and the CA from the perspective of target error bound. We propose to use this relation to improve the segmentation performance of class alignment, which is desired for dense prediction tasks. 
We provide the following theorem: 

\textbf{Theorem 2} If we assume there is a hypothesis space $\mathcal{H}$ for segmentation model $h^{\theta, \phi}$ and a hypothesis space $\mathcal{H}_D$ for domain classifiers $D^\psi$, and $\mathcal{H}\Delta\mathcal{H}\subset \mathcal{H}_D$, then we have
\begin{equation}
    \label{eq:bound_relation}
    \begin{aligned}
        \epsilon_T(h) &\leq \hat{\mathbb{UB}}_2 \leq \hat{\mathbb{UB}}_1,\\
        \hat{\mathbb{UB}}_1 &= \epsilon_S(h)+\frac{1}{2}d_{\mathcal{H}_D}(\mathcal{D}_S, \mathcal{D}_T)+\lambda,\\
        \hat{\mathbb{UB}}_2 &= \epsilon_S(h)+\frac{1}{2}d_{\mathcal{H}\Delta\mathcal{H}}(\mathcal{D}_S, \mathcal{D}_T)+\lambda.
    \end{aligned}
\end{equation}
The proof of this theorem is provided in Appendix.~\ref{sec:proof}.

Essentially, we limit the hypothesis space $\mathcal{H}$ and $\mathcal{H}_D$ in Eq.~(\ref{eq:bound_relation}) into the space of deep neural networks. Directly optimizing over $\hat{\mathbb{UB}_2}$ might be hard to converge since $\hat{\mathbb{UB}_2}$ is a tighter upper bound for the prediction error on target domain. 
The bounds relation in Eq.~(\ref{eq:bound_relation}) shows that the $\hat{\mathbb{UB}}_1$ is an upper bound of $\hat{\mathbb{UB}}_2$. This provides us a clue to improve the training process of class alignment, i.e.,  \textit{the domain alignment can be a global constraint and narrow down the searching space for the class alignment}. This also implies that integrating the domain alignment and class alignment might boost the training efficiency as well as the prediction performance of UDA. An ideal training process is illustrated in Fig.~\ref{fig:optim} where the searching space of $\hat{\mathbb{UB}}_2$~(CA) is constantly bounded by that of $\hat{\mathbb{UB}}_1$~(DA), ensuring the whole training process converge stably. This inspires us to design a new model, and we are explaining next in details about our model structures, losses and training process. 

\subsection{CALI Structure}
\label{sec:structure}
The existing DA or CA works usually involve a bilateral game. In CA, the game is between a feature extractor and a family of classifiers. The two players are optimized over the discrepancy of the two classifiers (note here the two players are the two classifiers vs. the feature extractor) in an opposite manner. In DA, the game happens between a segmentation net and a domain discriminator. The two players are optimized over the domain discrimination in an opposite way. It has been empirically showed \cite{vu2019advent, tsai2018learning} that DA performs well if the domain alignment happens to the prediction probability (after \CommentSty{Softmax()}). However, according to the identified relation in Eq.~(\ref{eq:bound_relation}), the two upper bounds $\hat{\mathbb{UB}}_1$ and $\hat{\mathbb{UB}}_2$ need to use the same feature, hence we connect the domain alignment and class alignment using a shared feature extractor and propose a novel concept called PTGS (see Fig.~\ref{fig:trilateral}) to illustrate an interesting structure to integrate DA and CA. Both $C_s$ and $D$ have game with $G$, but there is no game between $C_s$ and $D$, hence we call this game as \textit{pseudo-trilateral game}. Furthermore, as defined in Eq.~(\ref{eq:hdh_distance}), $h$ and $h^{'}$ are two different hypotheses, thus we have to ensure the classifiers in $C_s$ are different during the training.

\begin{figure}
{
\centering
  {\includegraphics[width=\linewidth]{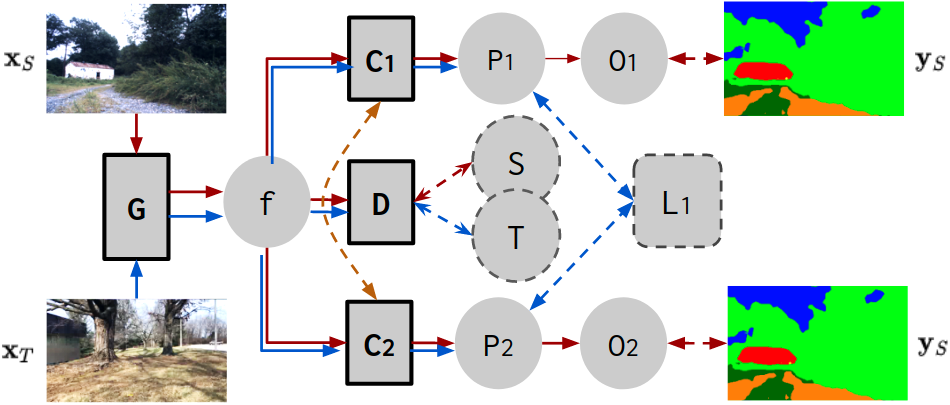} 
  } 
\caption{\small CALI network structure. See Section. \ref{sec:structure} for more details.}  \vspace{-10pt}
\label{fig:net} 
}
\end{figure}

Following the concept of PTGS, we design the structure of our CALI model as shown in Fig.~\ref{fig:net}. Four networks are involved, a shared feature extractor $G$, a domain discriminator $D$ and two classifiers $C_1$ and $C_2$. $f$ represents the shared features; $P_1/O_1$ and $P_2/O_2$ are the probability/class predictions for $C_1$ and $C_2$, respectively; $S/T$ represent the source domain label (1) and target domain label (0); and $L_1$ represents the $L_1$ distance measure between two probability distributions. The one-way solid arrows indicate the forward propagation of the data flow while the two-way dashed arrows indicate losses are generated. The red arrows represent the source-related data while the blue ones represent the target-related data. The orange two-way dashed line indicates the structural regularization loss between the $C_1$ and $C_2$.

\subsection{CALI Losses}
We denote raw images from source or target domain as $\mathbf{x}$, and the label from source domain as $\mathbf{y}$. We use semantic labels in source domain to train all of the nets, but the domain discriminator, in a supervised way, see the solid red one-way arrow in Fig.~\ref{fig:net}. We need to minimize the supervised segmentation loss since Eq.~(\ref{eq:bound_relation}) and other related Eqs suggest that the source prediction error is also part of the upper bound of target error.  The supervised segmentation loss for training CALI is defined as 
\begin{equation}
    \label{eq:seg_loss}
    \begin{aligned}
    &\mathcal{L}_{seg} (G, C_1, C_2) = \frac{1}{2}\Bigg( \Bigg. \mathbb{E}_{(\mathbf{x}, \mathbf{y})\sim \mathcal{D}_S} \left[ -\mathbf{y}\log (C_1(G(\mathbf{x}))) \right] + \\
    &~~~~~~~~~~~~~~~~~~~~~~~~~~~\mathbb{E}_ {(\mathbf{x}, \mathbf{y})\sim \mathcal{D}_S} \left[ -\mathbf{y}\log (C_2(G(\mathbf{x}))) \right]\Bigg. \Bigg)\\
    &= -\frac{1}{2}\mathbb{E}_{(\mathbf{x}, \mathbf{y})\sim \mathcal{D}_S} [\mathbf{y}_S \log \left( (C_1(G(\mathbf{x}))\odot (C_2(G(\mathbf{x})))) \right)],
    \end{aligned}
\end{equation}
where $\odot$ represents the element-wise multiplication between two tensors.

To perform domain alignment, we need to define the joint loss function for $G$ and $D$ 

\begin{equation}
    \label{eq:domain_loss}
        \mathcal{V}_1(G, D) = -\left(  \mathcal{CE}_S(\mathbf{x}) + \mathcal{CE}_T(\mathbf{x})\right),
\end{equation}
where no segmentation labels but domain labels are used, and we use the standard cross-entropy to compute the domain classification loss for both source ($\mathcal{CE}_S(\mathbf{x})$) and target data ($\mathcal{CE}_T(\mathbf{x})$). We have
\begin{equation}
    \label{eq:source_ce}
    \begin{aligned}
    \mathcal{CE}_S(\mathbf{x}) &= \mathbb{E}_{\mathbf{x}\sim \mathcal{D}_S} [\mathcal{CE}([1, 0]^T, [D(G(\mathbf{x})), 1-D(G(\mathbf{x}))]^T)]\\
    &= \mathbb{E}_{\mathbf{x}\sim \mathcal{D}_S} [-\log (D(G(\mathbf{x})))].
    \end{aligned}
\end{equation}
and
\begin{equation}
    \label{eq:target_ce}
    \begin{aligned}
    \mathcal{CE}_T(\mathbf{x}) &= \mathbb{E}_{\mathbf{x}\sim \mathcal{D}_T} [\mathcal{CE}([0, 1]^T, [D(G(\mathbf{x})), 1-D(G(\mathbf{x}))]^T)]\\
    &= \mathbb{E}_{\mathbf{x}\sim \mathcal{D}_T}[ -\log (1-D(G(\mathbf{x})))],
    \end{aligned}
\end{equation}
Note we include $G$ in Eq.~(\ref{eq:source_ce}) since both the source data and target data are passed through the feature extractor. This is different than standard GAN, where the real data is directly fed to $D$, without passing through the generator.

To perform class alignment, we need to define the joint loss function for $G$, $C_1$, and $C_2$
\begin{equation}
    \label{eq:class_loss}
    \begin{aligned}
    &\mathcal{V}_2(G, C_1, C_2) = \mathbb{E}_{\mathbf{x}\sim \mathcal{D}_T}\left [ d(C_1(G(\mathbf{x})), C_2(G(\mathbf{x}))) \right ],
    \end{aligned}
\end{equation}
where $d(\cdot, \cdot)$ is the distance measure between two distributions from the two classifiers. In this paper, we use the same $L_1$ distance in \cite{saito2018maximum} as the measure, thus $d(p, q) = \frac{1}{K}|p-q|_1$, where $p$ and $q$ are two distributions and $K$ is the number of label classes. 

To prevent $C_1$ and $C_2$ from converging to the same network throughout the training, we use the cosine similarity as a weight regularization to maximize the difference of the weights from $C_1$ and $C_2$, i.e.,  
\begin{equation}
    \label{eq:structural_loss}
    \mathcal{WR}(C_1, C_2)=\frac{\mathbf{w}_1\cdot \mathbf{w}_2}{\left\|\mathbf{w}_1 \right\|\left\|\mathbf{w}_2 \right\|},
\end{equation}
where $\mathbf{w}_1$ and $\mathbf{w}_2$ are the weight vectors of $C_1$ and $C_2$, respectively. 

\subsection{CALI Training}
We integrate the training processes of domain alignment and class alignment to systematically train our CALI model. To be consistent with  Eq.~(\ref{eq:bound_relation}), we adopt an iterative mechanism that alternates between domain alignment and class alignment. 
Our training process is pseudo-coded in Algorithm 1. 

\begin{algorithm}
\label{algo:training}
    \caption{\small CALI Training Process}
    {\small
     \textbf{Input:} Source dataset $\mathcal{U}_s$; Target dataset $\mathcal{U}_t$; Initial model $G, C_1, C_2$ and $D$; Maximum iterations $M$; Iteration interval $I$.\\
     \textbf{Output:} Updated model parameters $\phi_{G}, \theta_{C_1}, \theta_{C_2}$ and $\psi_{D}$\\
     \textbf{Initialization:} \textit{is\_domain=True; is\_class=False;}\\
        \For{m $\leftarrow$ 1~to~M}
            {   
                \If{$m \% I==0$ and $m \neq 0$}{
                    is\_domain = not is\_domain;\\
                    is\_class = not is\_class;\\
                  }
                \tcp{Eq.~(\ref{eq:seg_loss})}
                $\min_{\phi_{G}, \theta_{C_1}, \theta_{C_2}} \mathcal{L}_{seg}(G, C_1, C_2);$\\ 
                \tcp{Eq.~(\ref{eq:structural_loss})}
                $\min_{\theta_{C_1}, \theta_{C_2}} \mathcal{WR}(C_1, C_2);$\\
                \If{is\_domain}{
                    \tcp{Eq.~(\ref{eq:domain_loss})}
                    $\max_{\psi_{D}} \min_{\theta_{G}} \mathcal{V}_1(G, D);$\\
                }
                \If{is\_class}{
                    \tcp{Eq.~\ref{eq:class_loss}}
                    $\max_{\theta_{C_1}, \theta_{C_2}} \min_{\phi_{G}} \mathcal{V}_2(G, C_1, C_2);$\\
                }
            }
     Return $\phi_{G}$, $\theta_{C_1}$, $\theta_{C_2}$ and $\psi_{D}$;
    } 
\end{algorithm}

Note the adversarial training order of $\mathcal{V}_1$ in Algorithm 1 is $\max_{\psi_D} \min_{\phi_G}$, instead of the $\min_{\phi_G} \max_{\psi_D}$, meaning in each training iteration we first train the feature extractor and then the discriminator. The reason for this order is because we empirically find that the feature from $G$ is relatively easy for $D$ to discriminate, hence if we train $D$ first, then the $D$ might become an accurate discriminator in the early stage of training and there will be no adversarial signals for training $G$, thus making the whole training fail. The same order applies to training of the pair of $G$ and $Cs$ with $\mathcal{V}_2$.

\begin{figure*}
{
\centering
  {\includegraphics[width=\linewidth]{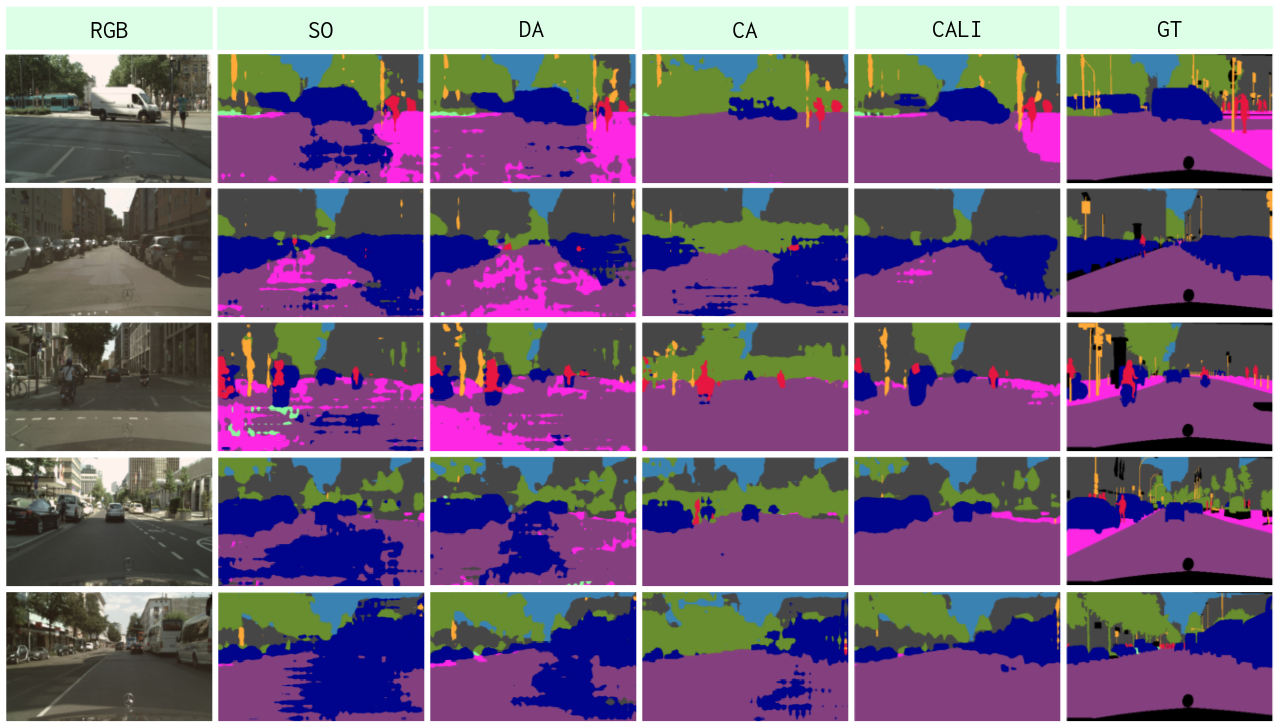} 
  } 
\caption{\small Qualitative results on adaptation GTA5$\rightarrow$Cityscapes. Results of our proposed model is listed in the last second column. GT represents the ground-truth labels.}
\label{fig:city_images} 
}
\end{figure*}

\subsection{Visual Planner}
We design a visual receding horizon planner to achieve feasible visual navigation by combining the learned image segmentation. 
Specifically, first we compute a library of motion primitives~\cite{howard2007optimal, howard2008state} $\mathcal{M} = \left \{ \mathbf{p}_1, \mathbf{p}_2, \cdots, \mathbf{p}_n \right \}$ 
where each $\mathbf{p}_* = \left \{ \mathbf{x}_1, \mathbf{x}_2, \cdots, \mathbf{x}_m \right \}$ is a single primitive. 
We use $\mathbf{x}_{*} = \begin{bmatrix}
x & y & \psi
\end{bmatrix}^T$ to denote a robot pose. 
Then we project the motion primitives to the image plane and compute the navigation cost function for each primitive based on the evaluation of collision risk in image space and target progress. Finally, we select the primitive with minimal cost  
to execute. The trajectory selection problem can be defined as:
\begin{equation}
    \label{eq:traj_selection}
    \mathbf{p}_{optimal} = \underset{\mathbf{p}}{\text{argmin}}~ w_1 \cdot C_{c}(\mathbf{p}) + w_2 \cdot C_{t}(\mathbf{p}),
\end{equation}
where $C_{c}(\mathbf{p}) = \sum_j^m c_c^j$ and $C_{t}(\mathbf{p}) = \sum_j^m c_t^j$ are the collision cost and target cost of one primitive $\mathbf{p}$, and $w_1$, $w_2$ are corresponding weights, respectively.

\begin{figure}
  \centering
  	\subfigure[]
  	{\label{fig:25}\includegraphics[width=0.3\linewidth]{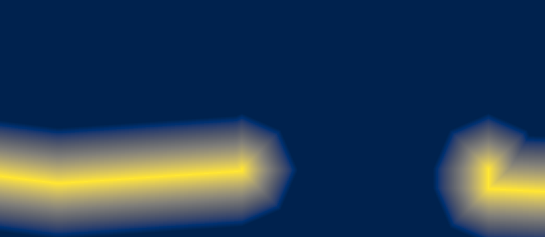}}
  	\subfigure[]
  	{\label{fig:55}\includegraphics[width=0.3\linewidth]{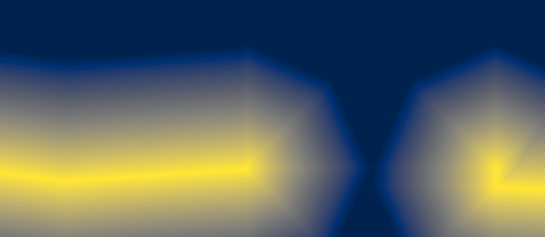}}
  	\subfigure[]
  	{\label{fig:100}\includegraphics[width=0.3\linewidth]{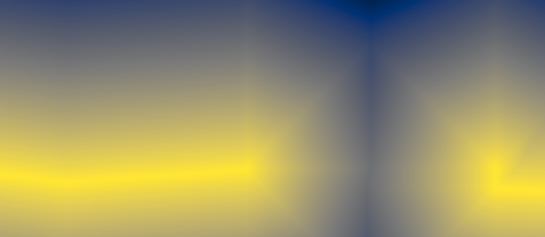}} \vspace{-10pt}
  \caption{\small Different SEDFs with varying scale factors of (a)~$\alpha=0.25$, (b)~$\alpha=0.55$ and (c)~$\alpha=1.00$. Values range from 0 to 1 by the color from blue to yellow.}\label{fig:sedf}  
\end{figure}

To efficiently evaluate the collision risk in the learned segmentation images, we first classify the classes in terms of their navigability, e.g., in off-road environments, grass and mulch are classified as navigable while tree and bush are classified as non-navigable. In this case, we are able to extract the boundary between the navigable space and the non-navigable space. We treat the boundary part close to the bottom line of the image as the \textit{obstacle boundary}. We further use the obstacle boundary to generate a Scaled Euclidean Distance Field (SEDF), where the values fall in $[0, 1]$, representing the risk level at the pixel position. Examples of different SEDF with different scale factors can be seen in Fig.~\ref{fig:sedf}.

Assume $\mathbf{x}^j$ is the $j^{th}$ pose in one primitive and its image coordinates are $\left ( u^j, v^j \right )$, then the collision risk for $\mathbf{x}^j$ is
\begin{equation}
    \label{eq:collision_cost}
        c_c^j = E[u^j, v^j],
\end{equation}
where $E$ represents the SEDF image.

\begin{figure*}
{
\centering
  {\includegraphics[width=\linewidth]{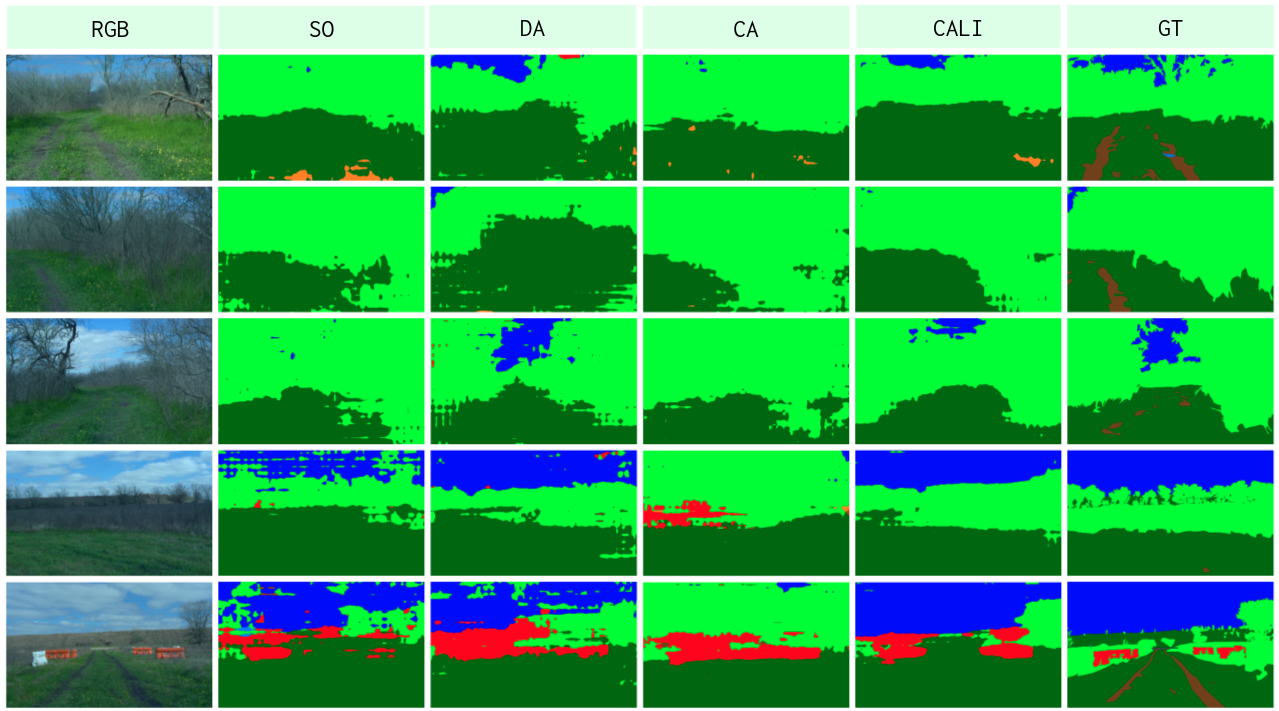} 
  } 
\caption{\small Qualitative results on adaptation RUGD$\rightarrow$RELLIS. Results of our proposed model is listed in the last second column. GT represents the ground-truth labels.} 
\label{fig:outdoor_images} 
}
\end{figure*}

To evaluate target progress during the navigation progress, we propose to use the distance on $SE(3)$ as the metric. We define three types of frames: world frame $F_w$, primitive pose frame $F_{pj}$, and goal frame $F_g$. The transformation of $F_{pj}$ in $F_w$ is denoted as $\mathbf{T}_{wpj}$ while that of $F_g$ in $F_w$ is $\mathbf{T}_{wg}$. A typical approach to represent the distance is to split a pose into a position and an orientation and define two distances on $\mathbb{R}^3$ and $SO(3)$. Then the two distances can be fused in a weighted manner with two strictly positive scaling factors $a$ and $b$ and with an exponent parameter $p\in [1, \infty]$~\cite{bregier2018defining}:
\begin{equation}
    \label{eq:se3_dist_1}
    \begin{aligned}
        d(\mathbf{T}_{wpj}, \mathbf{T}_{wg}) = \Bigg[ \Bigg. &a \cdot d_{rot}(\mathbf{R}_{wpj}, \mathbf{R}_{wg})^p+\\
        &b \cdot d_{trans}(\mathbf{t}_{wpj}, \mathbf{t}_{wg})^p\Bigg. \Bigg]^{1/p}.
    \end{aligned}
\end{equation}
We use the Euclidean distance as $d_{trans}(\mathbf{t}_{wpj}, \mathbf{t}_{wg})$, the Riemannian distance over $SO(3)$ as $d_{rot}(\mathbf{R}_{wpj}, \mathbf{R}_{wg})$ and set $p$ as $2$. Then the distance (target cost) between two transformation matrices can be defined~\cite{park1995distance} as:
\begin{equation}
\label{eq:se3_dist_2}
\begin{aligned}
    c_t^j &= d(\mathbf{T}_{wpj}, \mathbf{T}_{wg}) \\
          &= \left [ a \cdot \left \| \log(\mathbf{R}_{wpj}^{-1}\mathbf{R}_{wg}) \right \|^2+b \cdot \left \| \mathbf{t}_{wpj} - \mathbf{t}_{wg} \right \|^2 \right ]^{1/2}.
\end{aligned}
\end{equation}


\section{Experiments}
\label{sec:experiments}

\begin{figure*}
{
\centering
  {\includegraphics[width=1\linewidth]{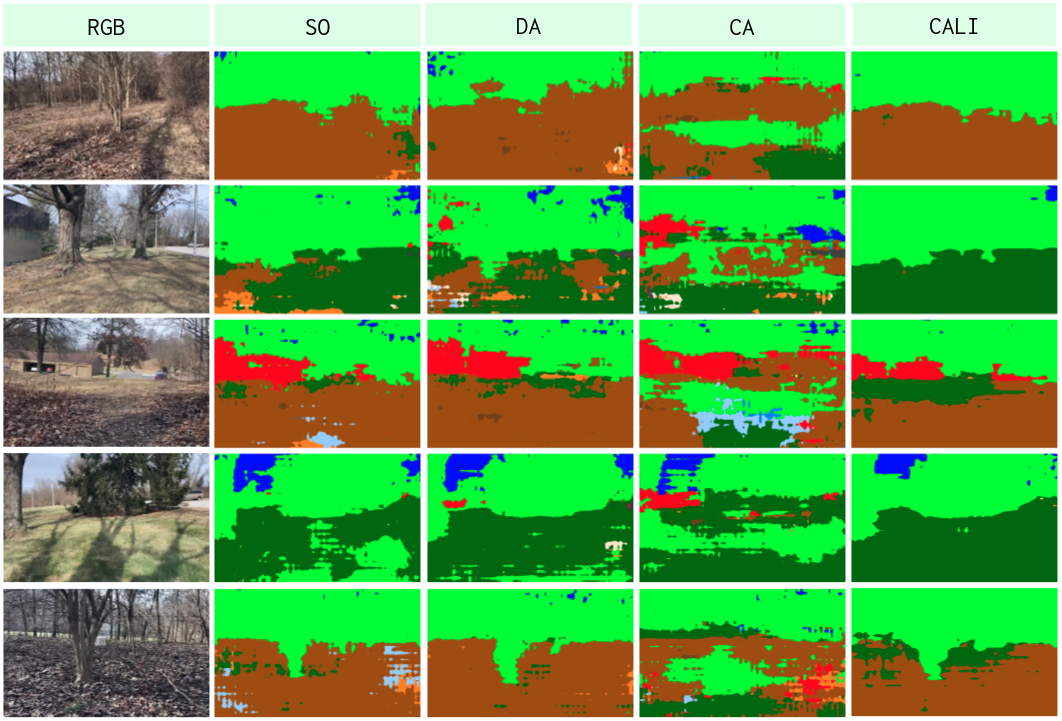} 
  } 
\caption{\small Qualitative results on adaptation RUGD$\rightarrow$MESH. Results of our proposed model is listed in the last column.
}  
\label{fig:mesh_images} 
}
\end{figure*}

\subsection{Datasets}
We evaluate CALI together with several baseline methods on a few challenging domain adaptation scenarios, where several public datasets, e.g., GTA5~\cite{richter2016playing}, Cityscapes~\cite{Cordts2016Cityscapes}, RUGD~\cite{RUGD2019IROS}, RELLIS~\cite{jiang2020rellis3d}, as well as a small self-collected dataset, named MESH (see the first column of Fig.~\ref{fig:mesh_images}), are investigated. 
The GTA5 dataset contains \num{24966} synthesized high-resolution images in the urban environments from a video game and pixel-wise semantic annotations of 33 classes. 
The Cityscapes dataset consists of \num{5000} finely annotated images whose label is given for 19 commonly seen categories in urban environments, e.g., road, sidewalk, tree, person, car, etc. 
The RUGD and RELLIS are two recently released datasets that aim to evaluate segmentation performance in off-road environments. 
The RUGD and the RELLIS contain 24 and 20 classes with \num{8000} and \num{6000} images, respectively. RUGD and RELLIS cover various scenes like trails, creeks, parks, villages, and puddle terrains. 
Our dataset, MESH, includes features like grass, trees (particularly challenging in winter due to foliage loss and monochromatic colors), mulch, etc. It helps us to further validate the performance of our proposed model for traversability prediction in challenging scenes, particularly the off-road environments.

{
\begin{table}
\caption{\small Quantitative comparison of different methods in UDA of GTA5$\rightarrow$Cityscapes. mIoU* represents the average mIoU over all of classes.} 
\centering 
{
\begin{tabularx}{0.7\linewidth}{ccccc}
\hline\hline 
  Class & SO & DA & CA & CALI \\[0.5ex]
\hline 
Road & 38.86 & 52.80 & \textbf{78.56} & 75.36 \\ [1ex]
Sidewalk & 17.47 & 18.95 & 2.79 & \textbf{27.12} \\ [1ex]
Building & 63.60 & 61.73 & 43.51 & \textbf{67.00} \\ [1ex]
Sky & 58.08 & 54.35 & 46.59 & \textbf{60.49} \\ [1ex]
Vegetation & 67.21 & 64.69 & 41.48 & \textbf{67.50} \\ [1ex]
Terrain & 7.63 & 7.04 & 8.37 & \textbf{9.56} \\ [1ex]
Person & \textbf{16.89} & 15.45 & 13.48 & 15.03\\ [1ex]
Car & 30.32 & 43.41 & 31.64 & \textbf{52.25} \\ [1ex]
Pole & 11.61 & \textbf{12.38} & 9.68 & 11.91 \\ [1ex]
mIoU* & 34.63 & 36.76 & 30.68 & \textbf{42.91} \\
\hline\hline 
\end{tabularx}\vspace{-10pt}
}
\label{tb:city_quantitative} 
\end{table}
}

\subsection{Implementation Details}
To be consistent with our theoretical analysis, the implementation of CALI only adopts the necessary indications by Eq.~(\ref{eq:bound_relation}). First, Eq.~(\ref{eq:bound_relation}) requires that the input of the two upper bounds (one for DA and the other one for CA) should be the same. Second, nothing else but only domain classification and hypotheses discrepancy are involved in Eq.~(\ref{eq:bound_relation}) and other related analyses (Eq.~(\ref{eq:theorem1}) - Eq.~(\ref{eq:ub2})). Accordingly, we strictly follow the guidance of our theoretical analyses. First, CALI performs DA in the  intermediate-feature level ($f$ in Fig.~\ref{fig:net}), instead of the output-feature level used in \cite{vu2019advent}. Second, we exclude the multiple additional tricks, e.g., entropy-based and multi-level features based alignment, and class-ratio priors in \cite{vu2019advent} and multi-steps training for feature extractor in \cite{saito2018maximum}. We also implement baseline methods without those techniques for a fair comparison. To avoid possible degraded performance bought by a class imbalance in the used datasets, we regroup those rare classes into classes with a higher pixel ratio. For example, we treat the building, wall, and fence as the same class; the person and rider as the same class in the adaptation of GTA5$\rightarrow$Cityscapes. In the adaptation of RUGD$\rightarrow$RELLIS, we treat the tree, bush, and log as the same class, and the rock and rockbed as the same class. Details about remapping can be seen in Fig.~\ref{fig:city_cloud} and Fig.~\ref{fig:outdoor_cloud} in Appendix.~\ref{sec:remapping}.

\begin{table}
\caption{\small Quantitative comparison of different methods in UDA of RUGD$\rightarrow$RELLIS. mIoU* is the average mIoU over all of classes.} 
\centering 
{
\begin{tabularx}{0.65\linewidth}{ccccc}
\hline\hline 
  Class & SO & DA & CA & CALI \\[0.5ex]
\hline 
Dirt & 0.00 & 0.53 & \textbf{3.23} & 0.01 \\ [1ex]
Grass & 64.78 & 61.63 & 65.35 & \textbf{67.08}\\ [1ex]
Tree & 40.79 & 45.93 & 41.51 & \textbf{55.80}\\ [1ex]
Sky & 45.07 & 67.00 & 2.31 & \textbf{72.99}\\ [1ex]
Building & 10.90 & \textbf{12.29} & 10.91 &10.28\\ [1ex]
mIoU* & 32.31 & 37.48 & 24.66 & \textbf{41.23}\\
\hline\hline 
\end{tabularx}
}
\label{tb:outdoor_quantitative} 
\end{table}

\begin{figure*} 
{
  \centering
    \subfigure[]
  	{\label{fig:city_comparison_1}\includegraphics[width=0.31\linewidth]{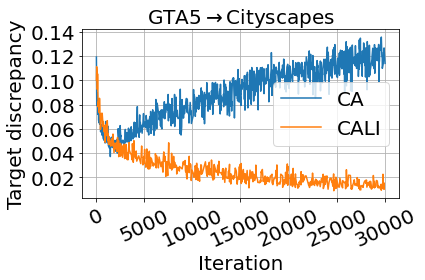}}
  	\subfigure[]
  	{\label{fig:city_comparison_2}\includegraphics[width=0.31\linewidth]{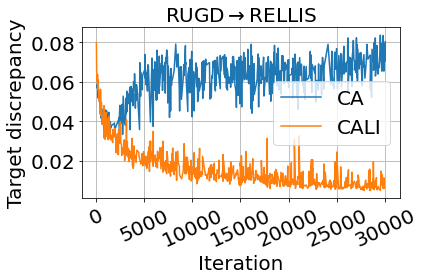}}  
  	\subfigure[]
  	{\label{fig:city_comparison_3}\includegraphics[width=0.31\linewidth]{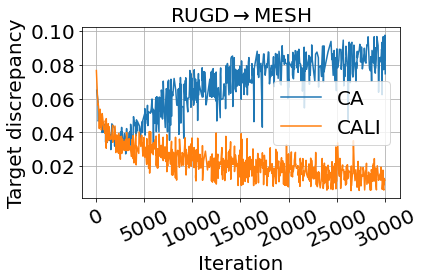}}  
  \caption{\small Target discrepancy changes during training process of (a) GTA5$\rightarrow$Cityscapes; (b) RUGD$\rightarrow$RELLIS; and (c) RUGD$\rightarrow$MESH.
  }
\label{fig:dis_comparison}  
}
\end{figure*}

\begin{figure} 
{
  \centering
    \subfigure[]
  	{\label{fig:d_outputs}\includegraphics[width=0.48\linewidth]{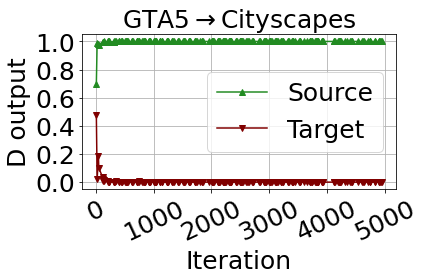}}
  	\subfigure[]
  	{\label{fig:minmax_dis}\includegraphics[width=0.48\linewidth]{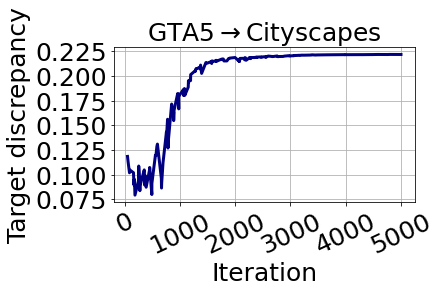}} 
  \caption{\small Using minmax can cause the collapse of training.
  }
\label{fig:minmax_curves}  
}
\end{figure}

\begin{figure}
{
\centering
  {\includegraphics[width=\linewidth]{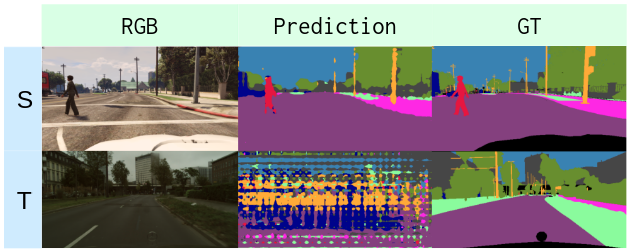} 
  } 
\caption{\small An example of collapsed trained model using minmax. }  \vspace{-10pt}
\label{fig:minmax_failure} 
}
\end{figure}

We use the PyTorch~\cite{paszke2019pytorch} framework for implementation.  Training images from source and target domains are cropped to be half of their original image dimensions. The batch size is set to 1 and the weights of all batch normalization layers are fixed. We use the ResNet-101~\cite{he2016deep} pretrained on ImageNet~\cite{deng2009imagenet} as the model $G$ for extracting features. We use the ASPP module in DeepLab-V2~\cite{chen2017deeplab} as the structure for $C_1$ and $C_2$. We use the similar structure in \cite{radford2015unsupervised} as the discriminator $D$, which consists of 5 convolution layers with kernel $4\times 4$ and with channel size $\left\{ 64, 128, 256, 512, 1 \right\}$ and stride of 2. Each convolution layer is followed by a Leaky-ReLU~\cite{maas2013rectifier} parameterized by 0.2, but only the last convolution layer is follwed by a Sigmoid function. During the training, we use SGD~\cite{bottou2010large} as the optimizer for $G, C_1$ and $C_2$ with a momentum of 0.9, and use Adam~\cite{kingma2014adam} to optimize $D$ with $\beta_1=0.9, \beta_2=0.99$. We set all SGD optimizers a weight decay of $5\text{e-}4$. The initial learning rates of all SGDs for performing domain alignment are set to $2.5\text{e-}4$ and the one of Adam is set as $1\text{e-}4$. For class alignment, the initial learning rate of SGDs is set to $1\text{e-}3$. All of the learning rates are decayed by a poly learning rate policy, where the initial learning rate is multiplied by $(1-\frac{iter}{max\_iters})^{power}$ with $power=0.9$. All experiments are conducted on a single Nvidia Geforce RTX 2080 Super GPU.

\subsection{Comparative Studies}

We present comparative experimental results of our proposed model, CALI, compared to different baseline methods -- Source-Only (SO) method, Domain-Alignment (DA)~\cite{vu2019advent} method, and Class-Alignment~\cite{saito2018maximum} method. Specifically, we first perform evaluations on a sim2real UDA in city-like environments, where the source domain is represented by GTA5 while the target domain is the Cityscapes. Then we consider a transfer of real2real in forest environments, where the source domain and target domain are set as RUGD and RELLIS, respectively. All models are trained with full access to the images and labels in the source domain and with only access to the images in the target domain. The labels in target datasets are only used for evaluation purposes. Finally, we further validate our model performance for adapting from RUGD to our self-collected dataset MESH. 

To ensure a fair comparison, all the methods use the same feature extractor $G$; both DA and CALI have the same domain discriminator $D$; both CA and CALI have the same two classifiers $C_1$ and $C_2$. We also use the same optimizers and optimization-related hyperparameters if any is used for models under comparison. 

We use the mean of Intersection over Union (mIoU) as the metric to evaluate each class and overall segmentation performance on testing images. IoU is computed as $\frac{n_{tp}}{n_{tp} + n_{fp} + n_{fn}}$, where $n_{tp}, n_{tn}, n_{fp}$ and $n_{fn}$ are true positive, true negative, false positive and false negative, respectively.

\subsubsection{GTA5$\rightarrow$Cityscapes}
Quantitative comparison results of GTA5$\rightarrow$Cityscapes are shown in Table.~\ref{tb:city_quantitative}, where segmentations are evaluated on 9 classes (as regrouped in Fig.~\ref{fig:city_cloud}). Our proposed method has significant advantages over multiple baseline methods for most categories and overall performance (mIoU*). 

\begin{figure*}
{
\centering
  {\includegraphics[width=\linewidth]{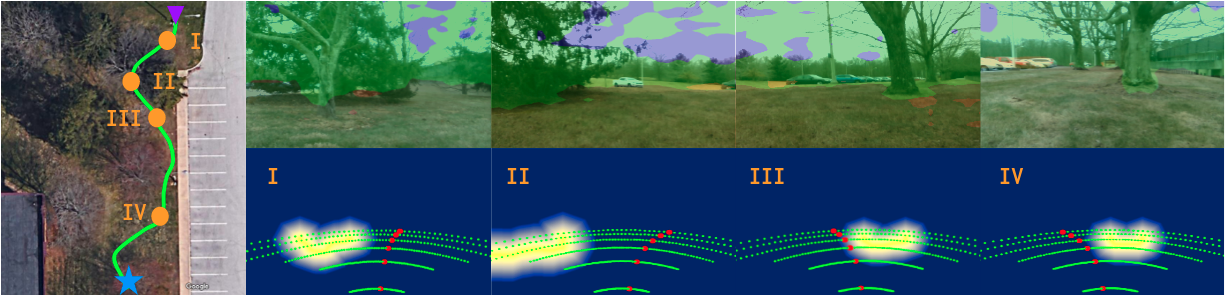} 
  } 
\caption{\small Navigation behaviors in MESH$\#1$ environment. The left-most column: top-down view of the environment; Purple triangle: the starting point; Blue star: the target point; We also show the segmentation (top row) and planning results (bottom row) at four different moments during the navigation, as shown from the second column to the last one. }  
\label{fig:nav_1} 
}
\end{figure*}

\begin{figure*}
{
\centering
  {\includegraphics[width=\linewidth]{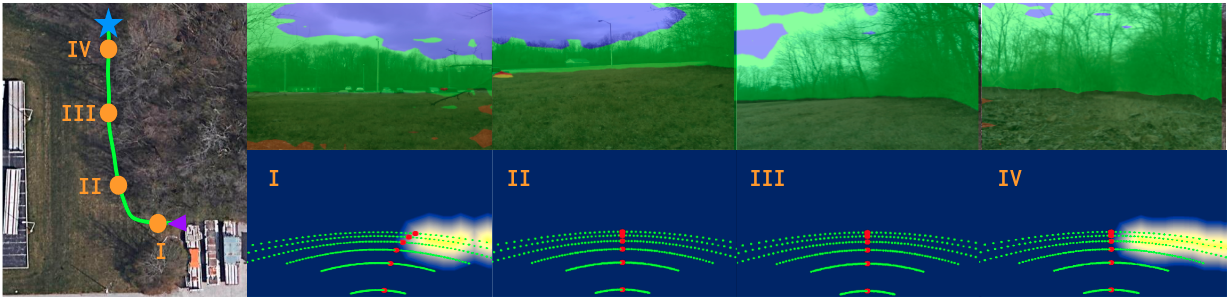} 
  } 
\caption{\small Navigation behaviors in MESH$\#2$ environment. Same legends with Fig.~\ref{fig:nav_1}.} 
\label{fig:nav_2} 
}
\end{figure*}

In our testing case, SO achieves the highest score for the class person even without any domain adaptation. One possible reason for this is the deep features of the source person and the target person from the model solely trained on source domain, are already well-aligned. If we try to interfere this well-aligned relation using \textit{unnecessary} additional efforts, the target prediction error might be increased (see the mIoU values of the person from the other three methods). We call this phenomenon as \textit{negative transfer}, which also happens to other classes if we compare SO and DA/CA, e.g., sidewalk, building, sky, vegetation, and so on. In contrast, CALI maintains an improved performance compared to either SO or DA/CA. We validate our analytical method for DA and CA (Section.~\ref{sec:bounds_relation}) by a comparison between CALI and baselines. This indicates either single DA or CA is problematic for semantic segmentation, particularly when we strictly follow what the theory supports and do not include any other training tricks (that might increase the training complexity and make the training unstable). This implies that integration of DA and CA is beneficial to each other with significant improvements, and more importantly, CALI is well theoretically supported, and the training process is easy and stable.

Fig.~\ref{fig:city_images} shows the examples of qualitative comparison for UDA of GTA5$\rightarrow$Cityscapes. We find that CALI prediction is less noisy compared to the baselines methods as shown in the second and third columns (sidewalk or car on-road), and shows better completeness (part of the car is missing, see the fourth column).

\subsubsection{RUGD$\rightarrow$RELLIS}
We show quantitative results of RUGD$\rightarrow$RELLIS in Table.~\ref{tb:outdoor_quantitative}, where only 5 classes\footnote{This is because other classes (in Fig.~\ref{fig:outdoor_cloud}) frequently appearing in source domain (RUGD) are extremely rare in target domain (RELLIS), hence no prediction for those classes occurs especially considering the domain shift.} are evaluated. It shows the same trend as Table.~\ref{tb:city_quantitative}. Both tables show that CA has the negative transfer (compared with SO) issue for either sim2real or real2real UDA. However, if we constrain the training of CA with DA, as in our proposed model, then the performance will be remarkably improved. Some qualitative results are shown in Fig.~\ref{fig:outdoor_images}.

\subsubsection{RUGD$\rightarrow$MESH} 
Our MESH dataset contains only unlabeled images that restrict us to show only a qualitative comparison for the UDA of RUGD$\rightarrow$MESH, as shown in Fig.~\ref{fig:mesh_images}. We have collected data in winter forest environments, which are significantly different than the images in the source domain (RUGD) - collected in a different season, e.g., summer or spring. These cross-season scenarios make the prediction more challenging. However, it is more practical to evaluate the UDA performance of cross-season scenarios, as we might have to deploy our robot at any time, even with extreme weather conditions, but our available datasets might be far from covering every season and every weather condition. From Fig.~\ref{fig:mesh_images}, we can still see the obvious advantages of our proposed CALI model over other baselines.


\subsection{Discussions}
\label{sec:ablation_studies}
In this section, we aim to discuss our model behaviors in more details. Specifically, first we will explain the advantages of CALI over CA from the perspective of training process. Second, we will show the vital influence of mistakenly using wrong order of adversarial training.

The most important part in CA is the discrepancy between the two classifiers, which is the only training force for the functionality of CA. It has been empirically studied in \cite{saito2018maximum} that the target prediction accuracy will increase as the target discrepancy is decreasing, hence the discrepancy is also an indicator showing if the training is on the right track. We compare the target discrepancy changes of CALI and our baseline CA in Fig.~\ref{fig:dis_comparison}, where the curves for the three UDA scenarios are presented from (a) to (c) and we only show the data before iteration 30k. It can be seen that before around iteration 2k, the target discrepancy of both CALI and CA are drastically decreasing, but thereafter, the discrepancy of CA starts to increase. On the other hand, if we impose a DA constraint over the same CA (iteratively), leading to our proposed CALI, then the target discrepancy will be decreasing as expected. This validates that integrating DA and CA will make the training process of CA more stable, thus improving the target prediction accuracy.

As mentioned in Algorithm 1, 
we have to use adversarial training order of $\max_{\psi_D} \min_{\phi_G}$, instead of $\min_{\phi_G} \max_{\psi_D}$. The reason for this is related to our designed net structure. Following the guidance of Eq.~(\ref{eq:bound_relation}), we use the same input to the two classifiers and the domain discriminator, hence the discriminator has to receive the intermediate-level feature as the input. If we use the order of $\min_{\phi_G} \max_{\psi_D}$ in CALI, then the outputs of the discriminator will be like Fig.~\ref{fig:d_outputs}, where the domain discriminator of CALI will quickly converge to the optimal state and it can accurately discriminate if the feature is from source or target domain. In this case, the adversarial loss for updating the feature extractor will be near 0, hence the whole training fails, which is validated by changes of the target discrepancy curve, as shown in Fig.~\ref{fig:minmax_dis}, where the discrepancy value is decreasing in a small amount in the first few iterations and then quickly increase to a high level that shows the training is divergent and the model is collapsed. This is also verified by the prediction results at (and after) around iteration 1k, as shown in Fig.~\ref{fig:minmax_failure}, where the first row is the source images while the second row is the target images.

\subsection{Navigation Missions}
To further show the effectiveness of our proposed model for real deployments,
we build a navigation system by combining the proposed CALI (trained with RUGD$\rightarrow$MESH set-up) segmentation model with our visual planner. We test behaviors of our navigation system in two different forest environments (named MESH$\#1$ in Fig.~\ref{fig:nav_1} and MESH$\#2$ in Fig.~\ref{fig:nav_2}), where our navigation system shows high reliability. 
In navigation tasks, the image resolution is $[400, 300]$, and the inference time for pure segmentation inference is around $33$ frame per second (FPS). However, since a complete perception system requires several post-processing steps, such as navigability definition, noise filtering, Scaled Euclidean Distance Field computation, motion primitive evaluation and so on, the response time for the whole perception pipeline (in python) is around $8$ FPS without any engineering optimization. The inference of segmentation for navigation is performed on an Nvidia Tesla T4 GPU. We set the linear velocity as $0.3m/s$ and control the angular velocity to track the selected motion primitive. The path length is $32.26m$ in Fig.~\ref{fig:nav_1} and $28.63m$ in Fig.~\ref{fig:nav_2}. Although the motion speed is slow in navigation tasks, as a proof of concept and with a very basic motion planner, the system behavior is as expected, and we have validated that the proposed CALI model is able to well accomplish the navigation tasks in unstructured environments.

\section{Conclusion} 
\label{sec:conclusion}
We present CALI, a novel unsupervised domain adaptation model specifically designed for semantic segmentation, which requires fine-grained alignments in the level of class features. We carefully investigate the relationship between a coarse alignment and a fine alignment in theory. The theoretical analysis guides the design of the model structure, losses, and training process. We have validated that the coarse alignment can serve as a constraint to the fine alignment and integrating the two alignments can boost the UDA performance for segmentation. The resultant model shows significant advantages over baselines in various challenging UDA scenarios, e.g., sim2real and real2real. We also demonstrate the proposed segmentation model can be well integrated with our proposed visual planner to enable highly efficient navigation in off-road environments.

\bibliographystyle{plainnat}
\bibliography{references}

\section*{Appendix}
\subsection{Proof of \textbf{Theorem 2}}
\label{sec:proof}
For a hypothesis $h$,
\begin{equation}
    \label{eq:proof_theorem2}
    \begin{aligned}
    \epsilon_T(h) &\leq \epsilon_T(h^{*})+\epsilon_T(h, h^{*})\\
    &=\epsilon_T(h^{*})+\epsilon_S(h, h^{*})-\epsilon_S(h, h^{*})+\epsilon_T(h, h^{*})\\
    &\leq \epsilon_T(h^{*})+\epsilon_S(h, h^{*})+|\epsilon_T(h, h^{*}) - \epsilon_S(h, h^{*})|\\
    &\leq \epsilon_T(h^{*})+\epsilon_S(h, h^{*})+\frac{1}{2}d_{\mathcal{H}\Delta\mathcal{H}}(\mathcal{D}_S, \mathcal{D}_T)\\
    &\leq \epsilon_T(h^{*})+\epsilon_S(h)+\epsilon_S(h^{*})+\frac{1}{2}d_{\mathcal{H}\Delta\mathcal{H}}(\mathcal{D}_S, \mathcal{D}_T)\\
    &= \epsilon_S(h)+\frac{1}{2}d_{\mathcal{H}\Delta\mathcal{H}}(\mathcal{D}_S, \mathcal{D}_T)+\epsilon_S(h^{*})+\epsilon_T(h^{*})\\
    &= \epsilon_S(h)+\frac{1}{2}d_{\mathcal{H}\Delta\mathcal{H}}(\mathcal{D}_S, \mathcal{D}_T)+\lambda\\
    &=\epsilon_S(h)+\sup_{h, h^{'}\in \mathcal{H}}|\text{P}_{\mathbf{x}\sim \mathcal{D}_S}\left [ h(\mathbf{x})\neq h^{'}(\mathbf{x}) \right ] -\\ &~~~~~~~~~~~~~~~~~~~~~~~\text{P}_{\mathbf{x}\sim \mathcal{D}_T}\left [ h(\mathbf{x})\neq h^{'}(\mathbf{x}) \right ]|+\lambda\\
    &=\epsilon_S(h)+\sup_{g\in \mathcal{H}\Delta\mathcal{H}}|\text{P}_{\mathbf{x}\sim \mathcal{D}_S}\left [ g(\mathbf{x})=1) \right ] -\\ &~~~~~~~~~~~~~~~~~~~~~~~\text{P}_{\mathbf{x}\sim \mathcal{D}_T}\left [ g(\mathbf{x})=1 \right ]|+\lambda\\
    &= \epsilon_S(h)+\sup_{g\in \mathcal{H}\Delta\mathcal{H}}| \text{P}_{\mathbf{x}\sim \mathcal{D}_S}\left [ g(\mathbf{x})=1) \right ] +\\ &~~~~~~~~~~~~~~~~~~~~~~~\text{P}_{\mathbf{x}\sim \mathcal{D}_T}\left [ g(\mathbf{x})=0 \right ]-1|+\lambda\\
    &\leq\epsilon_S(h)+\sup_{g\in \mathcal{H}\Delta\mathcal{H}}| \text{P}_{\mathbf{x}\sim \mathcal{D}_S}\left [ g(\mathbf{x})=1) \right ] +\\ &~~~~~~~~~~~~~~~\text{P}_{\mathbf{x}\sim \mathcal{D}_T}\left [ g(\mathbf{x})=0 \right ]| - \inf_{g\in \mathcal{H}\Delta\mathcal{H}} 1+\lambda\\
    &=\epsilon_S(h)+\sup_{g\in \mathcal{H}\Delta\mathcal{H}}| \text{P}_{\mathbf{x}\sim \mathcal{D}_S}\left [ g(\mathbf{x})=1) \right ] +\\ &~~~~~~~~~~~~~~~\text{P}_{\mathbf{x}\sim \mathcal{D}_T}\left [ g(\mathbf{x})=0 \right ]|+\lambda-1\\
    &=\epsilon_S(h)+\frac{1}{2}d_{\mathcal{H}}(\mathcal{D}_S, \mathcal{D}_T)+1+\lambda-1\\
    &=\epsilon_S(h)+\frac{1}{2}d_{\mathcal{H}}(\mathcal{D}_S, \mathcal{D}_T)+\lambda,
    \end{aligned}
\end{equation}
where $\lambda=\epsilon_S(h^{*})+\epsilon_T(h^{*})$ and $h^{*}$ is the ideal joint hypothesis (see the \textbf{Definition 2} in Section 4.2 of \cite{ben2010theory}).

We have the $4^{th}$, and the $8^{th}$ line because of the \textbf{Lemma 3} \cite{ben2010theory}; the $5^{th}$ line because of the \textbf{Theorem 2} \cite{ben2010theory}; the last second line because of the \textbf{Lemma 2} \cite{ben2010theory}. We have the $11^{th}$ line because $\sup|f_1 - f_2|= \sup f_1-\inf f_2 \leq \sup |f_1| - \inf f_2.~~~~~~~~~~~~~~~~~~~~~~~~~~~~~~~~~~~~~~~~~~~~~~~~~~~~~~~~~~~~~~~~~~~\blacksquare$\\

\subsection{Remapping of Label Space}
\label{sec:remapping}
\begin{figure}[t]
{
\centering
  {\includegraphics[width=\linewidth]{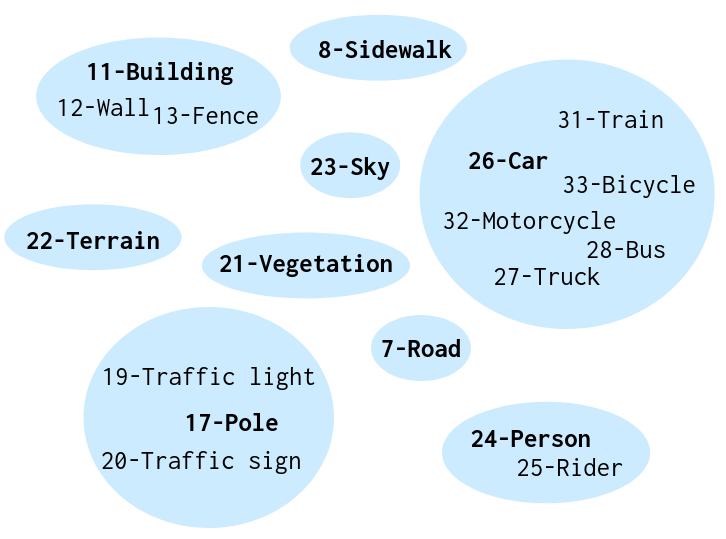} 
  } 
\caption{Lable remapping for GTA5$\rightarrow$Cityscapes. Name of each new group is marked as bold.}
\label{fig:city_cloud} 
}
\end{figure}

\begin{figure}[h]
{
\centering
  {\includegraphics[width=\linewidth]{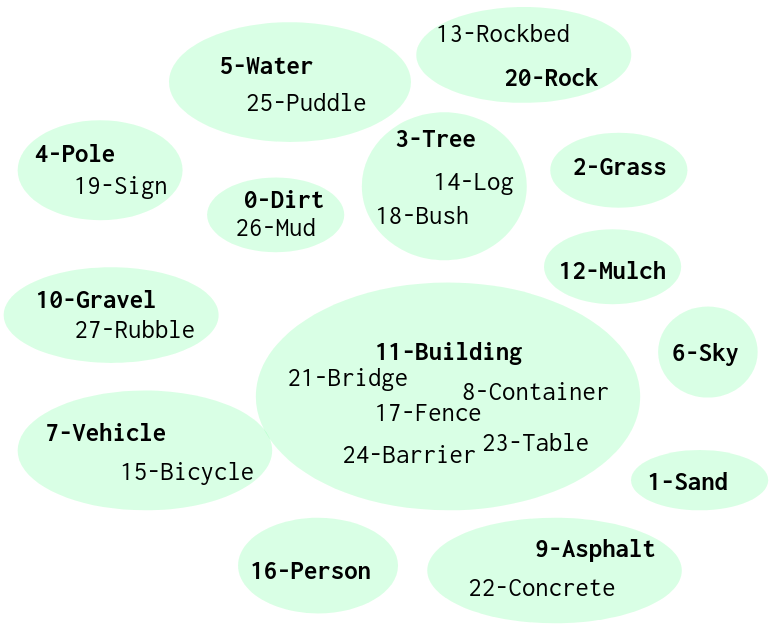} 
  } 
\caption{Lable remapping for RUGD$\rightarrow$RELLIS and RUGD$\rightarrow$MESH. Name of each new group is marked as bold.}  \vspace{-10pt}
\label{fig:outdoor_cloud} 
}
\end{figure}

We regroup the original label classes according to the semantic similarities among classes. In GTA5 and Cityscapes, we cluster the building, wall and fence as the same category; traffic light, traffic sign and pole as the same group; car, train. bicycle, motorcycle, bus and truck as the same class; and treat the person and rider as the same one. See Fig.~\ref{fig:city_cloud}. Similarly, we also have regroupings for classes in RUGD and RELLIS, as can be seen in Fig.~\ref{fig:outdoor_cloud}.

\end{document}